# Flying Triangulation –
# towards the 3D movie camera

**Florian Willomitzer, Svenja Ettl, Christian Faber, Gerd Häusler**

Institute of Optics, Information and Photonics,
Friedrich Alexander University Erlangen-Nuremberg,
Staudtstr. 7/B2, 91058 Erlangen, Germany

## 1 Introduction

The 3D acquisition of object shapes is of increasing importance. In the meantime, the market offers 3D-sensors for a wide spectrum of applications. Surprisingly, there are only a small number of approaches for a real-time 3D-camera. In this paper we will discuss, how the 3D measurement concept "Flying Triangulation" (FlyTri) [1] could be adapted (in the long term) to implement such a "3D movie camera". By this term, we understand a 3D-sensor that provides dense data, calculated from one single camera frame without exploiting lateral context information. Principally, FlyTri could provide this features, however, it has to be pushed further to its (information-theoretical) limits. FlyTri allows for a motion-robust real-time 3D acquisition of object surfaces using a hand-guided sensor. The data acquisition process is based on a multi-line light-sectioning approach. As a "single-shot principle", light sectioning enables the option to get 3D-data from one single camera exposure without exploiting neighbourhood information. Currently, our sensors project about 10 lines (each with 1000 pixels) per shot, reaching a significantly lower data efficiency than theoretically possible for a single-shot sensor. A high data density is then achieved by steadily moving the sensor around the object. We emphasize, that, due to fundamental information-theoretical reasons (incompressibility of space-time-bandwidth product), a single-shot sensor can never provide pixel-dense data if the object bandwidth is not severely limited. However, up to approximately 130 lines should be theoretically possible for a multi-line triangulation sensor with a 1000×1000 pixel camera (following from sampling theorem considerations) [2]. But there is a problem: if the line density is increased, severe indexing ambiguities occur. These ambiguities result in false 3D data which can harm or even disable the whole data acquisition process.

This contribution will discuss how to detect and correct the 3D data with a false index. The basic idea is to use yet unexploited information. We employ simultaneously acquired exposures from several cameras (instead of using the direct neighbourhood or a sequence of exposures from one camera). Compared to other approaches, our method does neither rely on spatial nor on temporal context information or the usage of an additional modality like colour.

## 2 Occurrence of ambiguities in the current indexing method

As soon as more than one single line is projected, any kind of light sectioning relies on indexing strategies. The first application of FlyTri was the intraoral measurement of human teeth. Hence, it was extremely important to develop an indexing strategy which is able to handle surface discontinuities. The main idea is to reserve a unique region on the camera chip for each line that directly corresponds to its index (see figure 1a). Inside the defined measurement depth $\Delta z$, each projected line $L1, ..., LN$ is located in an assigned region $A1, ..., AN$. Hence, surface points, measured inside $\Delta z$ by intersection with $L1, ..., LN$, are automatically imaged onto the corresponding areas $A1', ..., AN'$ on the camera chip. Here, the line index $n$ is determined. Together with the sub-pixel precise chip coordinates $(i,j)$ of the signal, $n$ encrypts the 3D coordinates $(x,y,z)$ of the object surface.

However, the method has a severe drawback: when a surface point is measured outside $\Delta z$, then, due to figure 1b, the signal is imaged outside of the correct area (here $A3'$). Hence, the signal is treated like generated from another projected line (here line $L3$). A false 3D point leads to an outlier in the final dataset. Figure 1c displays some examples. How to overcome this problem? A bigger measurement range $\Delta z$ could solve the problem. However, this is only possible if the number of projected lines will be reduced in order keep uniqueness inside $\Delta z$. Aiming for the 3D movie camera, our goal is the opposite: increasing the number of lines by simultaneously ensuring a comfortable large measurement volume. It is obvious that a new approach is necessary.

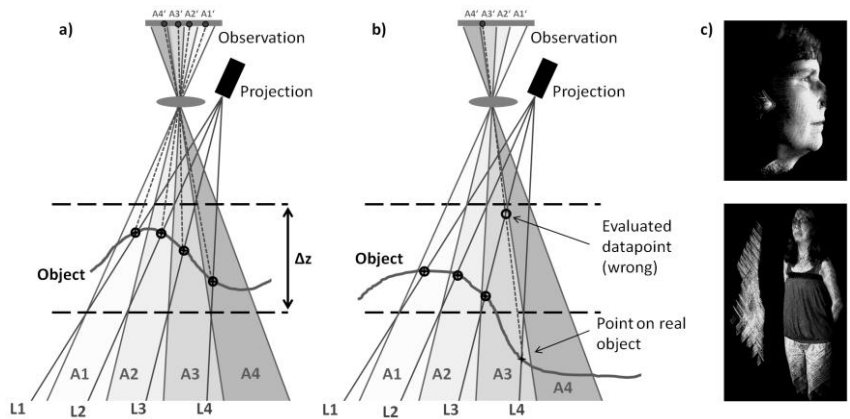

**Fig. 1.** a) Indexing approach of FlyTri (illustration for 4 lines): Surface points, measured inside $\Delta z$ by intersection of the projected lines are automatically imaged in their corresponding area onto the camera chip. b) False line indexing: the projected line L4 intersects the object surface outside the defined measurement depth. The signal coming from this point is imaged onto area A3' on the camera chip. Hence, it gets attached the wrong line index (3 instead of 4) which yields a wrong evaluated 3D datapoint. c) Examples for wrong 3D points caused by errors in the line indexing process.

## 3 Detection and correction of indexing ambiguities with the "Stereo-approach"

As mentioned above, the "stereo-approach" uses additional information from one or more additional cameras to get rid of false 3D data. We emphasize that the stereo approach is not related to active stereophotogrammetry, no lateral context information is used to identify corresponding points. In our FlyTri sensors, commonly a second camera for the acquisition of colour texture is included. Now, the main camera and the texture camera are used to acquire simultaneously images of the projected lines onto the object. After the 3D-evaluation from the main camera (data could be false indexed!), an index check with data from the texture camera is executed (see figure 2a): the evaluated 3D points are back-projected onto the chip of the texture camera. Here, the position of the back-projection is compared to the real signal onto the chip. If the signal and the back-projection match, the evaluated 3D point is correct. If not, it requires further processing as follows: a first (and not bad) approach is to simply delete false indexed points so that the registration cannot be disabled. Then only surface areas within the measurement range survive (see figure 1). If the number of lines will be largely increased, this range degenerates to a thin layer. Such a sensor is, practically, useless. However an algorithm for the correction of false indexed points is possible. A scheme for this algorithm is depicted in figure 2b.

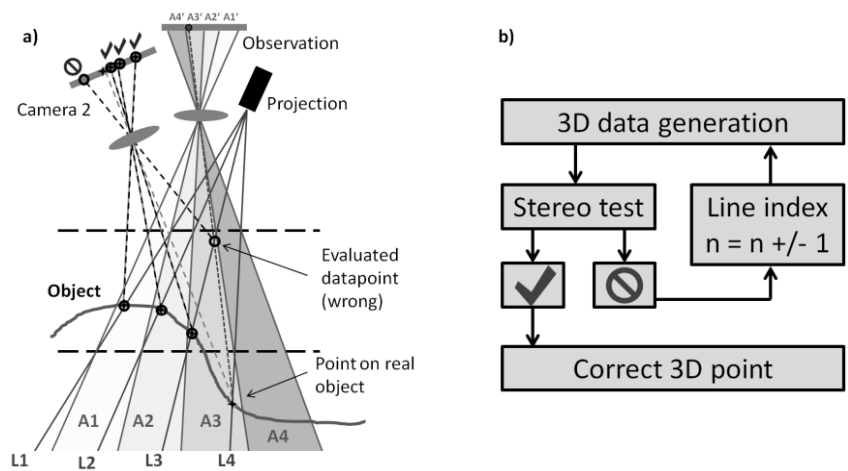

**Fig. 2.** a) The "stereo-test": Is a measured point correctly indexed? The evaluated data points are back projected onto the color camera chip. The position of the back projection is compared to the real signal on the chip. If the signal and the back-projection match, the evaluated 3D point is correctly indexed. b) Scheme for the correction algorithm.

## 3 Results and discussion

A first experiment was performed with the sparse 3D data set of a human body in front of a wall (the latter outside the measuring range) (see figure 3). From 1 Mio acquired 3D points (3a), 250.000 were identified as false and deleted (3b) or corrected (3c). In the final data sets, no remaining outliers are visible. However, there is a small possibility that a false point passes the stereo check. Several methods can prevent this: Reducing the cross section of lines and back-projections, placing the second camera in a "best case"-position or adding more cameras. All methods will be discussed in detail in further papers.

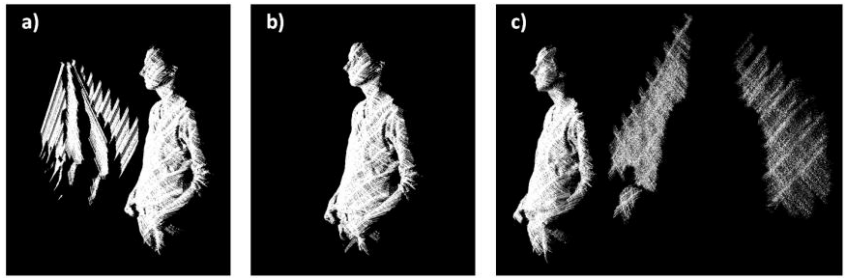

**Fig. 3.** First implementation: sparse measurement of a human body in front of a wall which is outside the measurement range. a) Raw data: 3D points of the wall are false. b) Indentified outliers deleted, no more outliers visible. c) Outliers corrected. The 3D data of the wall are correctly indexed.

The considerations about line density and limits provide an interesting consequence: if, due a large number of projected lines, the unique measurement depth approaches a thin layer, the task of generating a correct 3D point is not a task of 3D-evaluation anymore. It is only a task of indexing! A measurement principle that follows exactly this approach is the "Tomographic Triangulation" [2]. Under certain conditions, it is able to acquire pixel dense data within one single camera shot without using lateral context information and independent from ambient light. This view can be interpreted as an extension of the stereo approach with a large number of cameras.